\documentclass[conference]{IEEEtran}
\IEEEoverridecommandlockouts
\usepackage{cite}
\usepackage{amsmath,amssymb,amsfonts}
\usepackage{algorithmic}
\usepackage{graphicx}
\usepackage{textcomp}
\usepackage{xcolor}
\usepackage{booktabs}
\usepackage{cuted}
\usepackage{caption}
\usepackage{comment}
\usepackage{hyperref}

\def\BibTeX{{\rm B\kern-.05em{\sc i\kern-.025em b}\kern-.08em
    T\kern-.1667em\lower.7ex\hbox{E}\kern-.125emX}}

\begin{document}

\title{
Learning Sim-Grounded Policies for Bimanual Rope Manipulation from Human Teleoperation Data
}

\author{
Gina Wigginghaus$^{\star1}$, Tim Missal$^{\star1}$, Berk Guler$^{1,6}$, Simon Manschitz$^{6}$, Jan Peters$^{1-5}$

\thanks{$^\star$Equal contribution; $^1$Technical University of Darmstadt $^2$German Research Center for Artificial Intelligence (DFKI) $^3$Hessian.AI $^4$Robotics Institute Germany (RIG) $^5$Centre for Cognitive Science $^6$Honda Research Institute Europe GmbH; Corresponding author: \texttt{gina\_leonie.wigginghaus@stud.tu-darmstadt.de}}
\thanks{This work was supported by the German Federal Ministry of Research, Technology and Space (BMFTR) under the Robotics Institute Germany (RIG) and the Honda Research Institute EU, Germany.}}

\maketitle
\thispagestyle{empty}
\pagestyle{empty}

\begin{strip}
    \centering
    \vspace{-6em}
\includegraphics[width=\dimexpr\linewidth+3pt\relax]{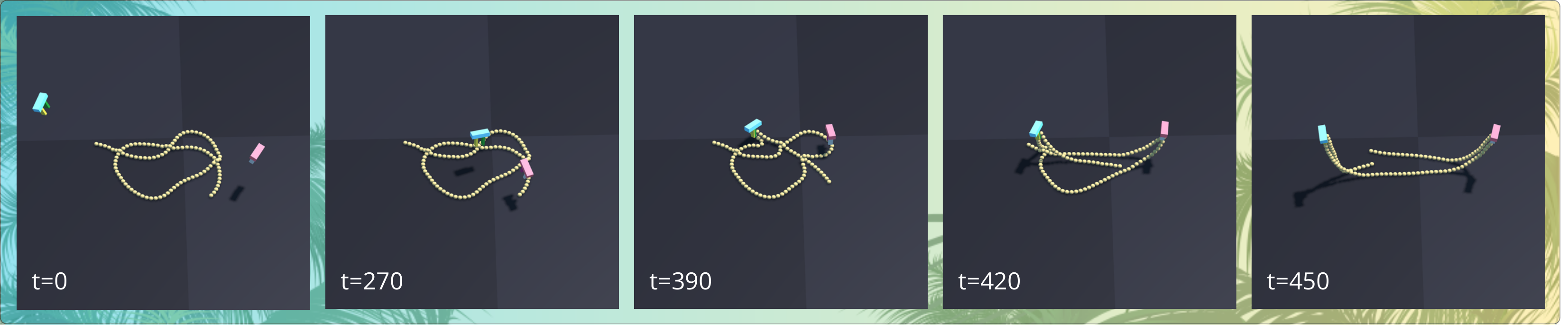}
    \vspace{-16pt}
    \captionof{figure}{\small
    Open-loop rollout of the particle-based ACT policy on an overhand-knotted rope excluded from the training set where $t$ indexes simulation frames at $30\,$Hz. The policy predicts a single macro action (grasp and pull) from the initial particle state at $t{=}0$; for overhand knot configurations like the one shown, a single well-placed pull is sufficient to fully resolve the knot. The policy successfully transfers this strategy to some unseen rope instances.}
    \label{fig:policy_rollout}
    \vspace{-5pt}
\end{strip}

\begin{abstract}
Deformable Linear Objects (DLOs) such as ropes and cables are widely encountered in both household and industrial applications, yet remain challenging to manipulate due to their infinite-dimensional configuration space and frequent self-occlusion. 
Imitation learning from teleoperation offers a practical path to bimanual DLO manipulation, but its scalability is limited by human effort, making the choice of observation space critical for generalization from small datasets. In this study, we investigate whether the lack of generalization in egocentric visual policies for the knot-untangling task stems from the observation space itself, rather than from the policy architecture or data scale. We compare two Action Chunking with Transformers policies trained on the same bimanual teleoperation data: a vision-based policy conditioned on two egocentric RGB streams from wrist-mounted cameras, and a state-based policy conditioned on the DLO's 3D particle state, extracted from an initial observation via multi-view fusion and evolved in a particle-based eXtended Position-Based Dynamics simulation. Evaluated open-loop on an unseen rope configuration, the state-based policy outperforms its visual counterpart with a $30.8\%$ reduction in L1 error when predicting the initial grasp-and-pull action, quantifying the observability gap between pixels and physics-consistent state, and pointing toward more data-efficient robot learning for the DLO manipulation task from limited human demonstrations.

\end{abstract}

\section{Introduction}

Tasks involving the manipulation of Deformable Linear Objects (DLOs), such as untangling ropes, remain one of the most challenging tasks in robotics due to high-dimensional state spaces and frequent self-occlusions \cite{grannen_untangling_2021, sundaresan_untangling_2021}. Traditional imitation learning approaches often attempt to map continuous RGB video streams from wrist-mounted cameras directly to robot actions~\cite{zhao_learning_2023, chi_diffusion_2024}. However, this ``end-to-end'' visual paradigm suffers from two major bottlenecks: computational inefficiency due to high-bandwidth data processing \cite{chen_piphen_2025}, and perceptual vulnerability when the robot’s own end-effectors (EEFs) occlude the object of interest \cite{zhu_viola_2023}.

In this work, we propose a shift from continuous visual tracking to a \textit{Snapshot-to-Action} paradigm grounded in physics-based simulation. Instead of requiring a persistent video stream, our method extracts the rope's topological state once, from a multi-view RGB-D observation fused across wrist-mounted cameras. This state is then mapped into a particle-based eXtended Position-Based Dynamics (XPBD) simulation using a Cosserat rod model~\cite{macklin_xpbd_2016,cosserat_theorie_1909, kugelstadt_position_2016, hsu_stable_2025}. This approach offers a dual advantage: it replaces high-dimensional visual inputs with a compact state representation, reducing compute and memory overhead and, more importantly, provides a physically consistent state representation that persists even when raw visual tracking fails.

We argue that for bimanual manipulation using end-effectors with wrist-mounted cameras, simulation is not merely an optional data source but a necessary state estimator. During manipulation, the robot arms inevitably occlude large segments of the rope, causing standard tracking algorithms to lose the trace of the rope \cite{xiang_trackdlo_2023}. By grounding the initial observation in an XPBD-based ``Digital Twin'', we create a representation anchor that maintains physical integrity where pixels provide no signal. 

We utilize Action Chunking with Transformers (ACT) to learn from human teleoperation demonstrations, collected without predefined action scripts or motion templates~\cite{zhao_learning_2023}. By focusing on the initial manipulation action, we benchmark the policy’s ability to interpret complex knots from a single observation state. 

To isolate the role of the observation space itself from confounders such as policy architecture or training data volume, we frame our evaluation as a controlled ablation: two policies with identical ACT backbones and identical training trajectories, differing only in whether they condition on raw egocentric pixels or on a physics-consistent particle state. We perform zero-shot evaluation on a different rope that does not appear in the training dataset, examining whether the observation space by itself dictates how effectively a policy generalizes across rope instances. Recent work on 3D representations has answered this question positively for rigid and articulated objects, but it remains unresolved for DLOs~\cite{ze_3d_2024, ze_generalizable_2025, zhu_learning_2023}.

Our main contributions and arguments are:

\begin{itemize}
    \item \textbf{Occlusion-Robust State Grounding:} We demonstrate that mapping a multi-view observation into a particle-based simulation provides a more robust representation for DLOs than raw pixels, specifically addressing the failure modes of visual tracking during bimanual occlusion. 
    \item \textbf{Observational Efficiency:} We show that a single multi-view state extraction, evolved in simulation, is sufficient for predicting complex human-like intent without continuous visual feedback, leading to reduced dependency on high-frequency visual feedback. We further show that inference time is reduced by 86.2\% and peak VRAM usage is reduced by 46.4\% using our data-efficient representation, enabling our approach to run on resource-constrained edge systems.

\end{itemize}

Through this controlled comparison, we quantify the observability gap between pixel-level observations and physics-consistent states in bimanual knot untangling.

\section{Related Work}

\textbf{Deformable Linear Object Manipulation.}
DLO manipulation is challenging due to infinite-dimensional configuration spaces, visual homogeneity, and frequent self-occlusion \cite{sanchez_robotic_2018,yin_modeling_2021,zhu_challenges_2021}. Recent work approaches rope untangling through topological state estimation and discrete action primitives. Grannen et al.~\cite{grannen_untangling_2021} predict keypoints for pull-and-pin primitives directly from RGB observations without reconstructing the full cable state. Sundaresan et al.~\cite{sundaresan_untangling_2021} refine this pipeline and extend it to non-planar knots by adding learned components for grasp refinement and recovery from failure modes, while other work generalizes the approach to dense multi-cable knots by planning over predicted topological graphs ~\cite{viswanath_autonomously_2022}. Recent assistive teleoperation work~\cite{guler_towards_2025} instead keeps the human in the loop as the planner. These methods share a common structure, where either perception produces a discrete state or a keypoint set that a planner executes using fixed motion primitives, or the human supplies topological reasoning at runtime. We instead learn continuous bimanual trajectories end-to-end from diverse limited human demonstrations, using a particle-based representation that captures finer geometric nuances than purely topological models and sidesteps the need to enumerate primitives at planning time.

\textbf{3D vs.\ 2D Observation Spaces for Imitation Learning.}
Recent work shows that structured 3D representations outperform raw 2D images when the training data is small. DP3 shows that diffusion policies conditioned on sparse point clouds generalize across space, viewpoint, appearance, and instance variations where equivalent RGB policies fail, requiring only 10 to 40 demonstrations per task~\cite{ze_3d_2024}. iDP3 extends this to egocentric 3D representations for humanoid manipulation and empirically demonstrates that image-based counterparts overfit to the training scenario~\cite{ze_generalizable_2025}. GROOT makes a related argument with object-centric 3D priors, showing that end-to-end visual policies are brittle to background, viewpoint, and instance changes that structured representations absorb~\cite{zhu_learning_2023}. These results are established primarily for rigid and articulated objects. We extend this approach to DLOs, where the relevant 3D structure is not a set of discrete entities but an ordered one-dimensional manifold, and where the gripper is the dominant source of occlusion.

\textbf{Imitation Learning with ACT.}
Imitation learning for bimanual tasks has been significantly advanced by ACT~\cite{zhao_learning_2023}, which uses a Conditional Variational Autoencoder (CVAE) Transformer architecture to predict synchronized action chunks from multi-view RGB observations. ACT and related methods usually rely on raw visual streams throughout the trajectory, making the policy’s effectiveness dependent on the quality and comprehensiveness of the visual observations during execution. We extend this framework to a \textit{Snapshot-to-Action} paradigm by grounding the ACT policy in an XPBD-based particle state initialized from a multi-view RGB-D observation, ensuring spatial equivariance and robustness to the occlusions inherent in bimanual manipulation.

\section{Methodology}
\begin{figure*}[t]
    \centering
    \includegraphics[width=\textwidth]{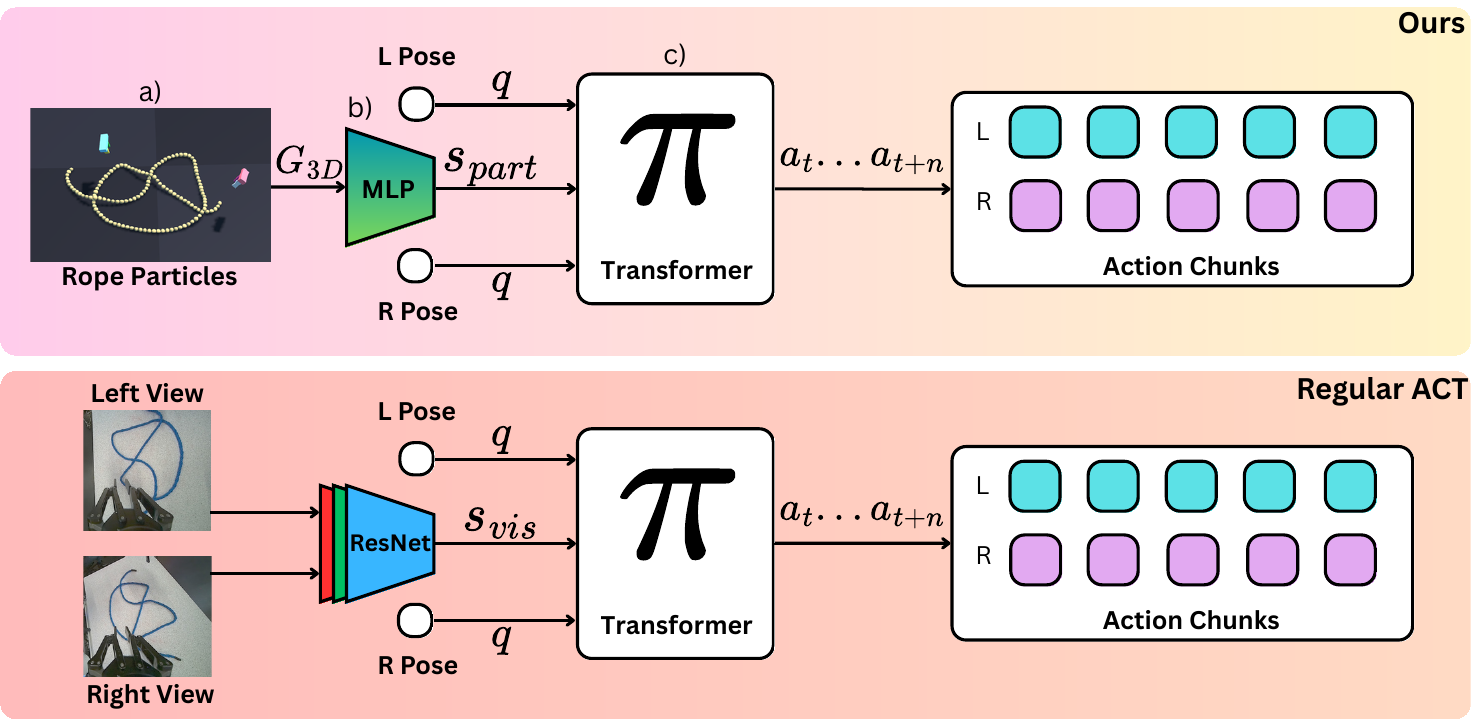}
    \caption{Comparative overview of our architecture and the regular ACT policy we are comparing against. a) Rope particle positions are extracted from an initial observation of the rope. b) The particle positions are encoded into a state vector through an MLP. c) A transformer model acts as the policy and predicts the next chunk of EEF movements.}
    \label{fig:architecture}

\end{figure*}

Our objective is to learn a policy $\pi$ that predicts a human-like manipulation action $a$ given an initial state $s$ of a tangled rope, comparing raw visual input and sim-grounded particle coordinates generated through physical re-execution.

\subsection{Data Acquisition and Bimanual Setup}
We utilize a dataset of human teleoperation demonstrations for the task of bimanual overhand-knot untangling, performed across four distinct ropes varying in length, color, and stiffness. The dataset comprises 96 demonstrations in which the human provides diverse, unconstrained manipulation strategies. Data was collected via a bimanual teleoperation interface, recording gripper poses $(x, y, z, q_x, q_y, q_z, q_w)$ and continuous gripper openness at $100$ Hz, resampled to $30$ Hz to match the simulation rate. Rather than filtering for ``perfect'' trajectories, we explicitly include sub-optimal demonstrations, treating the data as a diverse distribution of human topological reasoning that captures failed attempts and exploratory ``probing'' actions. Thus, we do not aim to train policies that fully untangle ropes. Instead, we focus on capturing human intent by predicting only the first macro action.

\subsection{Ground-Truth Generation via Simulation Playback}
To collect our dataset of particle states from teleoperation data, we employ a Simulation Playback pipeline:

\textbf{Snapshot Initialization:} Because the rope is heavily occluded by the robot's EEFs during manipulation, we temporarily extract multi-view RGB-D frames before the gripper initiates its grasp. Using the Segment Anything Model 2 (SAM2) \cite{ravi_sam_2024}, we generate binary masks of the rope and combine them with aligned depth maps, aggregating the views into a unified, voxel-downsampled 3D point cloud. To extract the rope's one-dimensional topology, we project this point cloud onto the 2D table plane and apply a Voronoi-based skeletonization to estimate its centerline~\cite{ravi_sam_2024,guler_towards_2025}. We then lift these keypoints back into a 3D ordered structure, resolving ambiguous self-intersections by evaluating the relative depth gradient $\nabla Z$ at crossing nodes. To standardize input dimensionality, we uniformly resample this centerline to a fixed sequence of $N{=}100$ particles via arc-length interpolation, producing the particle state $G_{3D} \in \mathbb{R}^{100 \times 3}$. This canonicalizes raw particle counts that ranged from 70 to 150 across rope instances, so each index corresponds to a consistent fractional position along the arc length.

\textbf{Action Re-execution and Labeling:} We spawn a simulated gripper in the XPBD environment, execute the human's recorded teleoperation actions $a_{\text{human}}(t)$, and record the resulting particle trajectory $\{G_{3D}(t)\}_{t=0}^{T}$ synchronized with those actions. This yields a continuous, fully observed state signal throughout the demonstration, including phases in which the gripper heavily occludes the rope and visual tracking would fail, producing the $(\text{state}, \text{action})$ pairs used to train the particle-based policy.

\subsection{Policy Learning with Action Chunking}
We build upon the ACT framework~\cite{zhao_learning_2023}. ACT models the policy $\pi$ as a CVAE. During training, a transformer encoder compresses the ground-truth action sequence into a latent style variable $z$, while a transformer decoder conditions on $z$, the current observation, and learned action queries to predict a chunk of $k$ future bimanual actions. At inference time, $z$ is sampled from the prior, and overlapping chunks are smoothed via temporal ensembling.

We define two input variants for the observation $s$, adapting the encoder accordingly while sharing the decoder architecture, action head, and CVAE latent structure. Both variants additionally receive the current bimanual proprioception $q \in \mathbb{R}^{16}$ as an embedded input token. For each arm, this consists of an 8-dimensional state: the end-effector pose (3D position and 4D quaternion) and a 1D scalar for gripper openness.

\textbf{Vision-based State ($s_{\mathrm{vis}}$):} Multi-view RGB images from both wrist-mounted cameras serve as input. A shared, pretrained ResNet-18 backbone independently extracts spatial feature maps from each view; these are projected to the transformer's hidden dimension ($d{=}512$) via a $1{\times}1$ convolution, concatenated along the spatial axis, and augmented with sinusoidal 2D position encodings.

\textbf{Particle-based State ($s_{\mathrm{part}}$):} To replace high-dimensional image processing with a structurally compact input, we introduce a custom ParticleEncoder that bypasses the ResNet backbone. The $N{=}100$ resampled 3D particle coordinates from the simulation playback are projected through a two-layer MLP with GELU activation and LayerNorm, producing one token per particle. Sinusoidal 1D positional encodings are added to preserve the sequential ordering along the rope's arc length.

\hyperref[fig:architecture]{Figure~\ref*{fig:architecture}} shows both architectures side by side. Apart from the encoder, both variants share the same backbone and are trained with the standard ACT loss combining L1 action reconstruction and KL regularization.

\section{Experiments}

The experimental setup consists of two 7-DoF Franka Panda arms with Robotiq 2F-140 grippers and wrist-mounted ZEDX stereo cameras. Demonstrations are collected via VR teleoperation (HTC Vive Pro) and provide 6-DoF EEF commands using VR controllers.
The 96 demonstrations are split into 64 training, 15 validation, and 17 test samples. The test set consists exclusively of demonstrations on the held-out rope, the stiffest of the four ropes and thus the most dynamically distinct from the training distribution. This evaluates zero-shot generalization to an unseen rope instance with unseen physical properties.

We quantitatively evaluate both policies through an offline comparison, using human teleoperation trajectories from the held-out test set as ground truth. For our particle-based policy, we extract an initial multi-view RGB-D snapshot to initialize the rope's geometry in simulation. To continuously provide the policy with updated particle state information, we propagate the simulation forward using Newton Physics, which employs an XPBD solver based on the Cosserat rod model \cite{the_newton_contributors_newton_2025,cosserat_theorie_1909}. Our implementation builds upon the rope simulation by Cai et al. and incorporates additional physical constraints from \cite{cai_gausstwin_2026,hsu_stable_2025}. In contrast, the baseline vision policy processes continuous multi-view RGB observations directly at each evaluation timestep.
Based on their respective state representations, both policies predict open-loop action chunks. We evaluate their ability to replicate human manipulation intent on unseen rope configurations using the L1 error, defined as the mean absolute distance between predicted and ground-truth bimanual EEF poses at the end of a $k{=}20$-step trajectory. 
\hyperref[table_results]{Table~\ref*{table_results} } 
reports these results alongside computational metrics evaluated on an NVIDIA RTX 4060 Ti GPU.

\begin{table}[h]
\centering
\caption{\textbf{Action prediction performance metrics.} Inference time is measured per single step, and Peak VRAM is the maximum PyTorch memory allocated during evaluation.}
\label{table_results}
\begin{tabular}{@{}lccc@{}}
\toprule
\textbf{Policy} & \textbf{L1 ($10^{-3}$) $\downarrow$} & \textbf{Inf. Time (ms) $\downarrow$} & \textbf{Peak VRAM (MB) $\downarrow$} \\ \midrule
Particles (S) & $\mathbf{9.70 \pm 7.95}$ & $\mathbf{2.95 \pm 0.05}$ & \textbf{334.40} \\
Vision (V)    & $14.02 \pm 8.44$         & $21.34 \pm 0.21$         & 623.66 \\ \bottomrule
\end{tabular}
\end{table}

To complement the quantitative evaluation, \hyperref[fig:policy_rollout]{Figure~\ref*{fig:policy_rollout}} shows an open-loop rollout of the particle-based policy on the held-out rope, which differs from the training instances in both appearance and stiffness. From the initial particle state alone, the policy predicts a grasp-and-pull action that resolves the overhand knot in a single macro step, transferring zero-shot to a rope instance with physical properties unseen during training. While this level of success is not achieved uniformly across all test samples, the rollout qualitatively demonstrates that the particle representation can encode sufficient information about the initial knot topology to plan a geometrically valid untangling strategy from a single snapshot.

\section{Discussion and Limitations}

While the particle-based representation offers significant advantages in terms of inference speed, memory footprint, and occlusion robustness, it introduces a reliance on the underlying physics model. Our current approach utilizes a fixed set of XPBD parameters for the Cosserat rod model and does not yet perform online system identification, so the simulation does not account for variations in rope stiffness, friction, or linear density. Any divergence in these parameters will eventually lead to drift between the simulated particle state and the physical rope. We envision two pathways to mitigate this: integrating an active System Identification module that estimates material properties from the initial snapshot, or a periodic re-grounding strategy that re-initializes the particle state from a fresh RGB-D snapshot after every grasp-and-pull action, effectively resetting tracking error between macro-actions.

Second, our vision baseline uses standard RGB ACT, while our particle pipeline implicitly exploits depth via multi-view reconstruction. The measured gap thus reflects the combined contribution of depth and explicit topological structure. Future work will evaluate an RGB-D ACT baseline to isolate the specific gains of depth versus structural representation.

Finally, raw RGB data remains inherently flexible, and vision-based policies are more intuitive and require no manual modeling of the object. However, their generalization scales with the size and diversity of the training data, so covering multiple rope types or changing environments demands extensive data collection. Our sim-grounded approach instead encodes physical priors directly into the state representation, offering a more sample-efficient path toward generalizable manipulation at the cost of requiring a reliable initial state estimator.

\section{Conclusion}

In this work, we investigated the impact of state representation on learning dexterous bimanual manipulation from limited and diverse human demonstrations. By benchmarking a vision-based ACT policy against a sim-grounded particle-based policy, we quantified the significant observability gap that exists when mapping raw pixels to complex deformable dynamics. Grounding a single visual snapshot into a particle representation yielded a $30.8\%$ reduction in L1 error and a 7.2x speedup in inference on a zero-shot rope instance. 

These gains suggest that for high-degree-of-freedom tasks like rope untangling, the bottleneck is not necessarily the amount of human data or the capacity of the transformer, but the structural clarity of the observation space. By shifting the burden of representation from end-to-end visual learning to physics-based grounding, we provide a path toward more data-efficient and computationally lean robot learning.

\bibliographystyle{IEEEtran}
\bibliography{references}

@inproceedings{guler_towards_2025,
	title = {Towards {Assistive} {Teleoperation} for {Knot} {Untangling}},
	booktitle = {German {Robotics} {Conference} ({GRC})},
	author = {Güler, Berk and Pompetzki, Kay and Manschitz, Simon and Peters, Jan},
	month = mar,
	year = {2025},
}

@misc{chi_diffusion_2024,
	title = {Diffusion {Policy}: {Visuomotor} {Policy} {Learning} via {Action} {Diffusion}},
	shorttitle = {Diffusion {Policy}},
	url = {http://arxiv.org/abs/2303.04137},
	doi = {10.48550/arXiv.2303.04137},
	urldate = {2026-04-24},
	publisher = {arXiv},
	author = {Chi, Cheng and Xu, Zhenjia and Feng, Siyuan and Cousineau, Eric and Du, Yilun and Burchfiel, Benjamin and Tedrake, Russ and Song, Shuran},
	month = mar,
	year = {2024},
	note = {arXiv:2303.04137 [cs]},
}

@misc{the_newton_contributors_newton_2025,
	title = {Newton: {GPU}-accelerated physics simulation for robotics and simulation research},
	copyright = {Apache-2.0},
	shorttitle = {Newton},
	url = {https://github.com/newton-physics/newton},
	urldate = {2026-04-16},
	author = {{The Newton Contributors}},
	month = apr,
	year = {2025},
}

@article{sanchez_robotic_2018,
	title = {Robotic {Manipulation} and {Sensing} of {Deformable} {Objects} in {Domestic} and {Industrial} {Applications}: {A} {Survey}},
	volume = {37},
	shorttitle = {Robotic {Manipulation} and {Sensing} of {Deformable} {Objects} in {Domestic} and {Industrial} {Applications}},
	url = {https://uca.hal.science/hal-01816189},
	doi = {10.1177/0278364918779698},
	number = {7},
	urldate = {2026-04-24},
	journal = {The International Journal of Robotics Research},
	publisher = {SAGE Publications},
	author = {Sanchez, Jose and Corrales Ramon, Juan Antonio and Bouzgarrou, Belhassen-Chedli and Mezouar, Youcef},
	month = jun,
	year = {2018},
	pages = {688 -- 716},
}

@misc{zhu_challenges_2021,
	title = {Challenges and {Outlook} in {Robotic} {Manipulation} of {Deformable} {Objects}},
	url = {http://arxiv.org/abs/2105.01767},
	doi = {10.48550/arXiv.2105.01767},
	urldate = {2026-04-24},
	publisher = {arXiv},
	author = {Zhu, Jihong and Cherubini, Andrea and Dune, Claire and Navarro-Alarcon, David and Alambeigi, Farshid and Berenson, Dmitry and Ficuciello, Fanny and Harada, Kensuke and Kober, Jens and Li, Xiang and Pan, Jia and Yuan, Wenzhen and Gienger, Michael},
	month = dec,
	year = {2021},
	note = {arXiv:2105.01767 [cs]},
}

@article{yin_modeling_2021,
	title = {Modeling, learning, perception, and control methods for deformable object manipulation},
	volume = {6},
	url = {https://www.science.org/doi/abs/10.1126/scirobotics.abd8803},
	doi = {10.1126/scirobotics.abd8803},
	number = {54},
	urldate = {2026-04-24},
	journal = {Science Robotics},
	publisher = {American Association for the Advancement of Science},
	author = {Yin, Hang and Varava, Anastasia and Kragic, Danica},
	month = may,
	year = {2021},
	pages = {eabd8803},
}

@misc{ravi_sam_2024,
	title = {{SAM} 2: {Segment} {Anything} in {Images} and {Videos}},
	shorttitle = {{SAM} 2},
	url = {http://arxiv.org/abs/2408.00714},
	doi = {10.48550/arXiv.2408.00714},
	urldate = {2026-04-24},
	publisher = {arXiv},
	author = {Ravi, Nikhila and Gabeur, Valentin and Hu, Yuan-Ting and Hu, Ronghang and Ryali, Chaitanya and Ma, Tengyu and Khedr, Haitham and Rädle, Roman and Rolland, Chloe and Gustafson, Laura and Mintun, Eric and Pan, Junting and Alwala, Kalyan Vasudev and Carion, Nicolas and Wu, Chao-Yuan and Girshick, Ross and Dollár, Piotr and Feichtenhofer, Christoph},
	month = oct,
	year = {2024},
	note = {arXiv:2408.00714 [cs]},
}

@misc{ze_generalizable_2025,
	title = {Generalizable {Humanoid} {Manipulation} with {3D} {Diffusion} {Policies}},
	url = {http://arxiv.org/abs/2410.10803},
	doi = {10.48550/arXiv.2410.10803},
	urldate = {2026-04-24},
	publisher = {arXiv},
	author = {Ze, Yanjie and Chen, Zixuan and Wang, Wenhao and Chen, Tianyi and He, Xialin and Yuan, Ying and Peng, Xue Bin and Wu, Jiajun},
	month = sep,
	year = {2025},
	note = {arXiv:2410.10803 [cs]},
}

@misc{ze_3d_2024,
	title = {{3D} {Diffusion} {Policy}: {Generalizable} {Visuomotor} {Policy} {Learning} via {Simple} {3D} {Representations}},
	shorttitle = {{3D} {Diffusion} {Policy}},
	url = {http://arxiv.org/abs/2403.03954},
	doi = {10.48550/arXiv.2403.03954},
	urldate = {2026-04-24},
	publisher = {arXiv},
	author = {Ze, Yanjie and Zhang, Gu and Zhang, Kangning and Hu, Chenyuan and Wang, Muhan and Xu, Huazhe},
	month = sep,
	year = {2024},
	note = {arXiv:2403.03954 [cs]},
}

@misc{zhu_learning_2023,
	title = {Learning {Generalizable} {Manipulation} {Policies} with {Object}-{Centric} {3D} {Representations}},
	url = {http://arxiv.org/abs/2310.14386},
	doi = {10.48550/arXiv.2310.14386},
	urldate = {2026-04-24},
	publisher = {arXiv},
	author = {Zhu, Yifeng and Jiang, Zhenyu and Stone, Peter and Zhu, Yuke},
	month = oct,
	year = {2023},
	note = {arXiv:2310.14386 [cs]},
}

@article{cosserat_theorie_1909,
	title = {Theorie des {Corps} deformables},
	volume = {81},
	copyright = {1909 Springer Nature Limited},
	issn = {1476-4687},
	url = {https://www.nature.com/articles/081067a0},
	doi = {10.1038/081067a0},
	language = {en},
	number = {2072},
	urldate = {2026-04-23},
	journal = {Nature},
	publisher = {Nature Publishing Group},
	author = {Cosserat, Eugène and Cosserat, François},
	month = jul,
	year = {1909},
	pages = {67--67},
}

@article{kugelstadt_position_2016,
	title = {Position and {Orientation} {Based} {Cosserat} {Rods}},
	language = {en},
	author = {Kugelstadt, T and Schömer, E},
	year = {2016},
}

@inproceedings{hsu_stable_2025,
	address = {New York, NY, USA},
	series = {{SIGGRAPH} {Conference} {Papers} '25},
	title = {Stable {Cosserat} {Rods}},
	isbn = {979-8-4007-1540-2},
	url = {https://dl.acm.org/doi/10.1145/3721238.3730618},
	doi = {10.1145/3721238.3730618},
	urldate = {2026-04-23},
	booktitle = {Proceedings of the {Special} {Interest} {Group} on {Computer} {Graphics} and {Interactive} {Techniques} {Conference} {Conference} {Papers}},
	publisher = {Association for Computing Machinery},
	author = {Hsu, Jerry and Wang, Tongtong and Wu, Kui and Yuksel, Cem},
	month = jul,
	year = {2025},
	pages = {1--10},
}

@inproceedings{macklin_xpbd_2016,
	address = {New York, NY, USA},
	series = {{MIG} '16},
	title = {{XPBD}: position-based simulation of compliant constrained dynamics},
	isbn = {978-1-4503-4592-7},
	shorttitle = {{XPBD}},
	url = {https://dl.acm.org/doi/10.1145/2994258.2994272},
	doi = {10.1145/2994258.2994272},
	urldate = {2026-04-23},
	booktitle = {Proceedings of the 9th {International} {Conference} on {Motion} in {Games}},
	publisher = {Association for Computing Machinery},
	author = {Macklin, Miles and Müller, Matthias and Chentanez, Nuttapong},
	month = oct,
	year = {2016},
	pages = {49--54},
}

@misc{zhu_viola_2023,
	title = {{VIOLA}: {Imitation} {Learning} for {Vision}-{Based} {Manipulation} with {Object} {Proposal} {Priors}},
	shorttitle = {{VIOLA}},
	url = {http://arxiv.org/abs/2210.11339},
	doi = {10.48550/arXiv.2210.11339},
	urldate = {2026-04-22},
	publisher = {arXiv},
	author = {Zhu, Yifeng and Joshi, Abhishek and Stone, Peter and Zhu, Yuke},
	month = mar,
	year = {2023},
	note = {arXiv:2210.11339 [cs]},
}

@misc{zhao_learning_2023,
	title = {Learning {Fine}-{Grained} {Bimanual} {Manipulation} with {Low}-{Cost} {Hardware}},
	url = {http://arxiv.org/abs/2304.13705},
	doi = {10.48550/arXiv.2304.13705},
	urldate = {2026-04-22},
	publisher = {arXiv},
	author = {Zhao, Tony Z. and Kumar, Vikash and Levine, Sergey and Finn, Chelsea},
	month = apr,
	year = {2023},
	note = {arXiv:2304.13705 [cs]},
}

@misc{chen_piphen_2025,
	title = {{PIPHEN}: {Physical} {Interaction} {Prediction} with {Hamiltonian} {Energy} {Networks}},
	shorttitle = {{PIPHEN}},
	url = {http://arxiv.org/abs/2511.16200},
	doi = {10.48550/arXiv.2511.16200},
	urldate = {2026-04-22},
	publisher = {arXiv},
	author = {Chen, Kewei and Long, Yayu and Shang, Mingsheng},
	month = nov,
	year = {2025},
	note = {arXiv:2511.16200 [cs]},
}

@misc{cai_gausstwin_2026,
	title = {{GaussTwin}: {Unified} {Simulation} and {Correction} with {Gaussian} {Splatting} for {Robotic} {Digital} {Twins}},
	shorttitle = {{GaussTwin}},
	url = {http://arxiv.org/abs/2603.05108},
	doi = {10.48550/arXiv.2603.05108},
	urldate = {2026-04-15},
	publisher = {arXiv},
	author = {Cai, Yichen and Jansonnie, Paul and Farias, Cristiana de and Arenz, Oleg and Peters, Jan},
	month = mar,
	year = {2026},
	note = {arXiv:2603.05108 [cs]},
}

@inproceedings{sundaresan_untangling_2021,
	title = {Untangling {Dense} {Non}-{Planar} {Knots} by {Learning} {Manipulation} {Features} and {Recovery} {Policies}},
	isbn = {978-0-9923747-7-8},
	url = {http://www.roboticsproceedings.org/rss17/p013.pdf},
	doi = {10.15607/RSS.2021.XVII.013},
	language = {en},
	urldate = {2026-02-27},
	booktitle = {Robotics: {Science} and {Systems} {XVII}},
	publisher = {Robotics: Science and Systems Foundation},
	author = {Sundaresan, Priya and Grannen, Jennifer and Thananjeyan, Brijen and Balakrishna, Ashwin and Ichnowski, Jeffrey and Novoseller, Ellen and Hwang, Minho and Laskey, Michael and Gonzalez, Joseph and Goldberg, Ken},
	month = jul,
	year = {2021},
}

@inproceedings{grannen_untangling_2021,
	title = {Untangling {Dense} {Knots} by {Learning} {Task}-{Relevant} {Keypoints}},
	issn = {2640-3498},
	url = {https://proceedings.mlr.press/v155/grannen21a.html},
	language = {en},
	urldate = {2026-02-27},
	booktitle = {Proceedings of the 2020 {Conference} on {Robot} {Learning}},
	publisher = {PMLR},
	author = {Grannen, Jennifer and Sundaresan, Priya and Thananjeyan, Brijen and Ichnowski, Jeffrey and Balakrishna, Ashwin and Viswanath, Vainavi and Laskey, Michael and Gonzalez, Joseph and Goldberg, Ken},
	month = oct,
	year = {2021},
	pages = {782--800},
}

@misc{viswanath_autonomously_2022,
	title = {Autonomously {Untangling} {Long} {Cables}},
	url = {http://arxiv.org/abs/2207.07813},
	doi = {10.48550/arXiv.2207.07813},
	urldate = {2026-02-27},
	publisher = {arXiv},
	author = {Viswanath, Vainavi and Shivakumar, Kaushik and Kerr, Justin and Thananjeyan, Brijen and Novoseller, Ellen and Ichnowski, Jeffrey and Escontrela, Alejandro and Laskey, Michael and Gonzalez, Joseph E. and Goldberg, Ken},
	month = jul,
	year = {2022},
	note = {arXiv:2207.07813 [cs]},
}

@article{xiang_trackdlo_2023,
	title = {{TrackDLO}: {Tracking} {Deformable} {Linear} {Objects} {Under} {Occlusion} {With} {Motion} {Coherence}},
	volume = {8},
	issn = {2377-3766},
	shorttitle = {{TrackDLO}},
	url = {https://ieeexplore.ieee.org/document/10214157},
	doi = {10.1109/LRA.2023.3303710},
	number = {10},
	urldate = {2026-02-22},
	journal = {IEEE Robotics and Automation Letters},
	author = {Xiang, Jingyi and Dinkel, Holly and Zhao, Harry and Gao, Naixiang and Coltin, Brian and Smith, Trey and Bretl, Timothy},
	month = oct,
	year = {2023},
	pages = {6179--6186},
}

\end{document}